\documentclass{easychair}

\usepackage{doc}

\usepackage{bm}
\usepackage{algorithm}
\usepackage{algorithmic}
\usepackage{amsmath,amssymb,graphicx}

\newcommand{\easychair}{\textsf{easychair}}

\newcommand{\vect}[1]{\bm{#1}}
\newcommand{\mat}[1]{\bm{#1}}
\title{Real-time Mixed-Integer Quadratic Programming for Driving Behavior-Inspired Speed Bump Optimal Trajectory Planning}
\author{
Van Nam Dinh \inst{1} \thanks{first and corresponding author},
Van Vy Phan \inst{1},
Thai Son Dang \inst{1},
Van Du Phan \inst{1},
The Anh Mai \inst{1},
Van Chuong Le \inst{1},
Sy Phuong Ho \inst{1},
Dinh Tu Duong \inst{1},
\and
Hung Cuong Ta \inst{1}
}

\institute{
  School of Engineering and Technology, Vinh University, Vinh, Vietnam\\
  \email{namdv@vinhuni.edu.vn}
 }

\authorrunning{Nam et al.}

\titlerunning{The {\easychair} Class File}

\begin{document}

\maketitle

\begin{abstract}
This paper proposes a novel methodology for trajectory planning in autonomous vehicles (AVs), addressing the complex challenge of negotiating speed bumps within a unified Mixed-Integer Quadratic Programming (MIQP) framework. By leveraging Model Predictive Control (MPC), we develop trajectories that optimize both the traversal of speed bumps and overall passenger comfort. A key contribution of this work is the formulation of speed bump handling constraints that closely emulate human driving behavior, seamlessly integrating these with broader road navigation requirements. Through extensive simulations in varied urban driving environments, we demonstrate the efficacy of our approach, highlighting its ability to ensure smooth speed transitions over speed bumps while maintaining computational efficiency suitable for real-time deployment. The method's capability to handle both static road features and dynamic constraints, alongside expert human driving, represents a significant step forward in trajectory planning for urban autonomous driving applications.
\end{abstract}

\setcounter{tocdepth}{2}
{\small
\tableofcontents}

%\section{To mention}
%
%Processing in EasyChair - number of pages.
%
%Examples of how EasyChair processes papers. Caveats (replacement of EC
%class, errors).

%------------------------------------------------------------------------------
\section{Introduction}
\label{sect:introduction}
\begin{figure}[tb]
	\begin{centering}
	\includegraphics[width=0.68\textwidth]{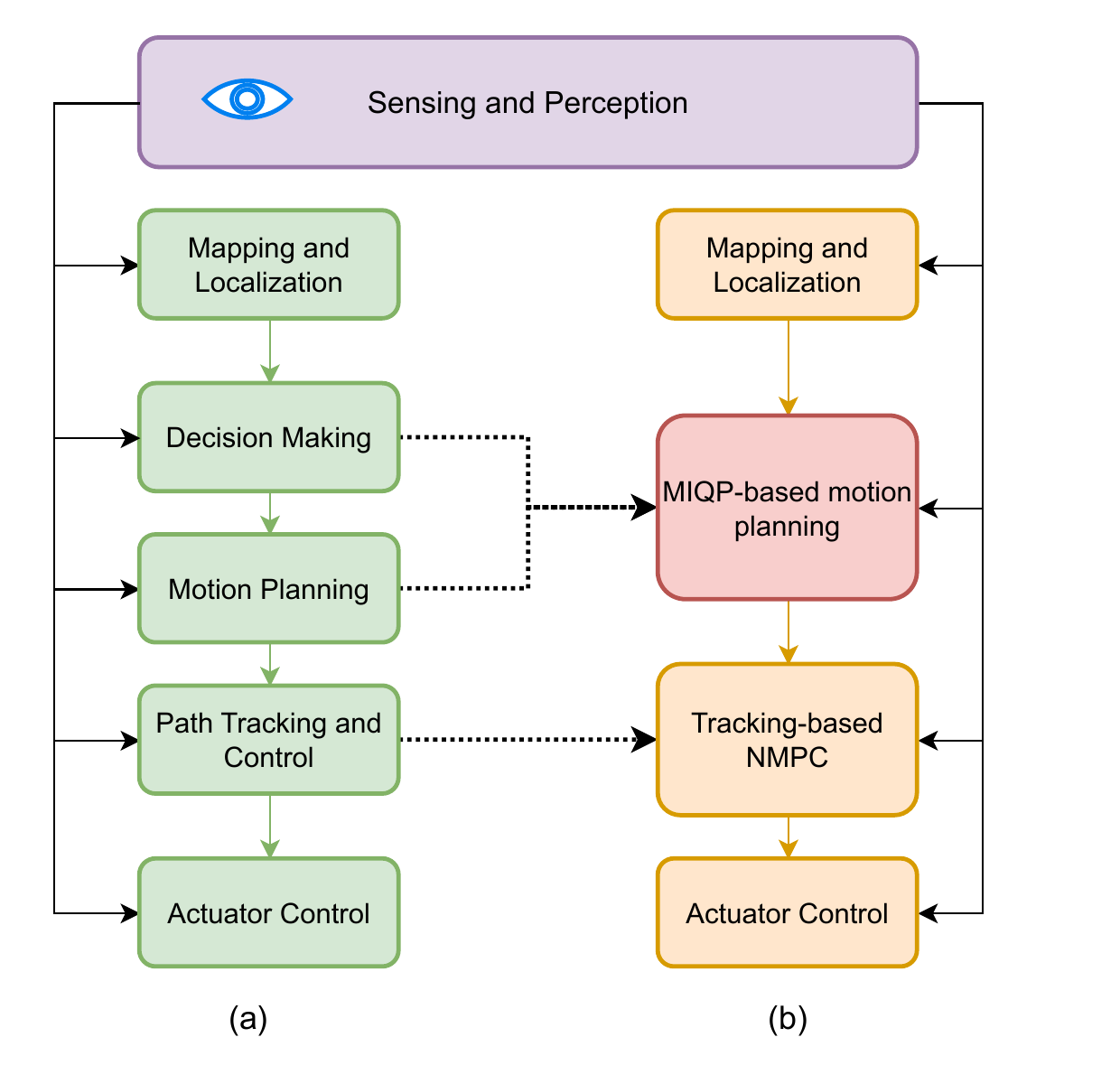}
	\caption{Motion planning control}
	\label{fig:auto_system}
	\end{centering}
\end{figure}

Autonomous vehicles (AVs) are poised to transform transportation, offering improvements in safety, efficiency, and accessibility \cite{paden2016survey, van2020hierarchical}. Figure~\ref{fig:auto_system} illustrates the architecture of an autonomous driving system, which is structured hierarchically and comprises several core components that work together to facilitate safe and efficient navigation, particularly in scenarios such as speed reduction zones. Figure~\ref{fig:auto_system}(a) outlines the system’s key modules: sensing and perception, mapping and localization, decision-making, motion planning, and control systems \cite{paden2016survey, van2020hierarchical}. Additionally, advanced techniques for perception and state estimation are critical \cite{van2022learning, van2023learning}.

In this study, we focus on the motion planning control for AVs \cite{paden2016survey}. Two major enhancements are introduced on the right side of the architecture, as shown in Fig.~\ref{fig:auto_system}(b). The first is the MIQP-based Motion Planning module, which employs Mixed-Integer Quadratic Programming to generate optimal trajectories. This is particularly useful for navigating complex scenarios with discrete decisions, such as speed bumps or construction zones \cite{qian2016optimal}. The second enhancement is a tracking-based NMPC module, utilizing Nonlinear Model Predictive Control to ensure robust tracking of the planned trajectory, while adapting in real time to changing conditions \cite{falcone2007predictive}.
By integrating these advanced motion planning and control techniques with high-level decision-making, this approach allows the autonomous system to handle complex driving situations with greater safety and efficiency, ensuring smooth operation across diverse driving conditions.

Navigating speed bumps is a critical yet often underappreciated component of trajectory planning for autonomous vehicles (AVs), especially in urban settings \cite{navarro2018real}. Speed bumps, which are traffic-calming measures, are implemented to slow vehicles down and enhance safety in areas with significant pedestrian or cyclist activity. While human drivers manage these obstacles with relative ease, they pose substantial challenges for autonomous systems due to the need to balance safety, ride comfort, and adherence to traffic regulations simultaneously \cite{navarro2018real, urban_av_challenges}. One of the core challenges lies in decelerating sufficiently to minimize impact while maintaining efficient travel times. Human drivers, particularly those with less experience, may brake too abruptly or fail to accelerate smoothly afterward, resulting in jerky motions that compromise passenger comfort. In contrast, autonomous systems have the potential to optimize these transitions through real-time trajectory adjustments, considering both the vehicle's dynamics and the surrounding environment.

Trajectory planning for AVs has been a major area of research, with a focus on ensuring safe and efficient navigation in complex, dynamic environments. Various strategies have been proposed, ranging from graph-based methods to continuous optimization techniques, each tackling different facets of the problem \cite{paden2016survey}. Dolgov et al. \cite{dolgov2010path} introduced the hybrid A* algorithm for path planning in semi-structured environments. Although this method handled scenarios with road boundaries and tight passages effectively, it did not explicitly account for dynamic obstacles or vehicle speed constraints, making it less suitable for navigating speed bumps or other irregular road conditions. Similarly, Ferguson et al. \cite{ferguson2008motion} developed an anytime D* search algorithm for urban driving, offering real-time path planning performance in environments with static obstacles. However, this approach did not address continuous dynamic constraints, limiting its ability to manage the speed modulation required for negotiating speed bumps.

More recently, Model Predictive Control (MPC) has gained attention as a robust tool for trajectory planning, owing to its capacity to incorporate complex vehicle dynamics and constraints into the optimization process. Li et al. \cite{li2015model} applied MPC to lane-change maneuvers, producing smooth, collision-free trajectories. Despite demonstrating the real-time effectiveness of MPC in trajectory generation, their work did not consider obstacle avoidance in dense environments or account for speed bump negotiation, which necessitates precise speed control to ensure passenger comfort.

Mixed-Integer Programming (MIP) has been widely employed to handle discrete decision-making in trajectory planning. Schouwenaars et al. \cite{schouwenaars2001mixed} used Mixed-Integer Linear Programming (MILP) to address aircraft collision avoidance. While this offered a robust framework for handling discrete constraints, the challenges posed by ground vehicle navigation—such as road boundaries, speed limits, and speed bumps—were not considered. Adapting MIP to AV trajectory planning introduces additional complexities, requiring the integration of both discrete obstacles (e.g., lane boundaries) and continuous variables (e.g., speed and position).

The issue of speed bump negotiation in autonomous driving has received limited attention. Handling speed bumps requires AVs to adjust their speed to ensure passenger comfort and protect the vehicle's suspension. This task becomes more complicated when speed bumps are combined with other constraints, such as dynamic obstacles or road boundaries. A study by Navarro et al. \cite{navarro2018real} explored using MPC to manage speed bumps in a parking lot setting; however, their approach focused on low-speed environments and did not address the challenges posed by speed bumps in more dynamic urban driving scenarios.

In this paper, we propose an advanced trajectory planning method that incorporates speed bump negotiation into the larger problem of autonomous driving in urban environments. While most existing methods for AV trajectory planning focus primarily on obstacle avoidance and lane-keeping, we specifically address the unique challenges posed by speed bumps. Our approach uses a Model Predictive Control (MPC) framework to generate optimal trajectories that balance speed regulation, comfort, and safety. By formulating the problem as a Mixed-Integer Quadratic Programming (MIQP) optimization, we can seamlessly integrate speed bump handling into the broader task of urban navigation.

The contributions of this paper are as follows:
\begin{itemize}
    \item A detailed analysis of vehicle dynamics related to speed bump traversal, highlighting the need for speed modulation to minimize discomfort and mechanical stress.
    \item A novel MIQP-based optimization framework that integrates speed bump negotiation into a unified trajectory planning solution for urban driving.
    \item Extensive simulations demonstrate the system’s ability to handle speed bumps smoothly while maintaining efficient, safe, and comfortable vehicle motion.
\end{itemize}

By focusing on speed bump negotiation within the context of AV trajectory planning, we address an important gap in the current literature. The proposed approach ensures that autonomous systems can handle the diverse range of road conditions found in urban environments, contributing to safer, more comfortable, and more efficient AV operations. The rest of this paper is organized as follows: Section \ref{sect:Methodology} presents our MIQP formulation and vehicle model.
Section \ref{sect:Results} discusses our MPC-based trajectory generation method. Section V presents our experimental setup and results, followed by a conclusion in Section VI.

%------------------------------------------------------------------------------
%------------------------------------------------------------------------------
\section{Methodology}
\label{sect:Methodology}
This work assumes that the autonomous vehicle's perception system can reliably detect speed bumps and provide accurate information about their positions within a predefined range. Specifically, the detection range is defined as $[d_{\text{min}}, d_{\text{max}}]$, where $d_{\text{min}}$ and $d_{\text{max}}$ represent the minimum and maximum distances from the vehicle at which speed bumps can be perceived.
\subsection{Vehicle Dynamics and Linearize Non-holonomic Constraints}
\subsubsection{Vehicle Dynamics}
Generally, the vehicle's dynamic model is a non-linear system \cite{van2020hierarchical, urban_av_challenges}, that can be represented as:
\begin{equation}
\dot{\mathbf{x}} = f(\mathbf{x}) = 
\begin{bmatrix}
\dot{x} \\
\dot{y} \\
\dot{\theta} \\
\dot{v}
\end{bmatrix}
=
\begin{bmatrix}
v \cos(\theta) \\
v \sin(\theta) \\
\frac{v}{L} \tan(\sigma) \\
a
\end{bmatrix}
\label{kinetic_model}
\end{equation}
where,
\( x \) and \( y \) are the coordinates of the vehicle's position,
\( \theta \) is the heading angle of the vehicle;
\( v \) is the linear velocity of the vehicle,
\( \sigma \) is the steering angle,
\( L \) is the wheelbase of the vehicle,
\( a \) is the acceleration of the vehicle.

However, nonlinear systems are challenging to handle in MIQP-MPC.
To enable real-time performance, we model the vehicle as a third-order linear point-mass system \cite{qian2016optimal}. The state vector $\vect{x}(k)$ and control input $\vect{u}(k)$ at time step $k$ are defined as:
\begin{equation}
\begin{aligned}
    \vect{x}(k) &= [x(k), v_x(k), a_x(k), y(k), v_y(k), a_y(k), \theta(k)]^\top \\
    \vect{u}(k) &= [j_x(k), j_y(k)]^\top
\end{aligned}
\end{equation}
where $(x, y)$ is the position, $(v_x, v_y)$ are velocities, $(a_x, a_y)$ are accelerations, $\theta$ is the heading angle, and $(j_x, j_y)$ are jerks in the longitudinal and lateral directions, respectively.

The discrete-time dynamics are given by:
\begin{equation}
\vect{x}(k+1) = \mat{A}\vect{x}(k) + \mat{B}\vect{u}(k)
\end{equation}
where
\begin{equation}
\mat{A} = \begin{bmatrix}
\mat{A}_d & \mat{0} \\
\mat{0} & \mat{A}_d
\end{bmatrix}, \quad
\mat{B} = \begin{bmatrix}
\mat{B}_d & \mat{0} \\
\mat{0} & \mat{B}_d
\end{bmatrix}
\end{equation}
with
\begin{equation}
\mat{A}_d = \begin{bmatrix}
1 & \Delta t & \frac{1}{2}\Delta t^2 \\
0 & 1 & \Delta t \\
0 & 0 & 1
\end{bmatrix}, \quad
\mat{B}_d = \begin{bmatrix}
\frac{1}{6}\Delta t^3 \\
\frac{1}{2}\Delta t^2 \\
\Delta t
\end{bmatrix}
\end{equation}
and $\Delta t$ is the time step.
The heading angle update can be approximated as:
\begin{equation}
\theta(k+1) = \theta(k) + \Delta t \cdot \frac{v_y(k)}{v_x(k)}
\end{equation}
The vehicle dynamics are modeled using a discrete-time kinematic model:
\begin{equation}
\left\{
\begin{aligned}
    x(k+1) &= x(k) + v_x(k)\Delta t + \frac{1}{2}a_x(k)\Delta t^2 + \frac{1}{6}j_x(k)\Delta t^3 \\
    y(k+1) &= y(k) + v_y(k)\Delta t + \frac{1}{2}a_y(k)\Delta t^2 + \frac{1}{6}j_y(k)\Delta t^3 \\
    v_x(k+1) &= v_x(k) + a_x(k)\Delta t + \frac{1}{2}j_x(k)\Delta t^2 \\
    v_y(k+1) &= v_y(k) + a_y(k)\Delta t + \frac{1}{2}j_y(k)\Delta t^2 \\
    a_x(k+1) &= a_x(k) + j_x(k)\Delta t \\
    a_y(k+1) &= a_y(k) + j_y(k)\Delta t
\end{aligned}
\right.
\end{equation}
where $\Delta t$ is the time step.

\subsubsection{Linearize Non-honolomic Constraints}
The lateral velocity \(v_y(k)\) is related to the longitudinal velocity \(v_x(k)\) and the steering angle \(\theta\) by:
\[
v_y = v_x \tan(\theta)
\]
Given the steering angle constraints \(\theta_\text{min}\) and \(\theta_\text{max}\), the lateral velocity \(v_y(k)\) must satisfy:
\[
v_y(k) \in [v_x(k) \tan(\theta_\text{min}), v_x(k) \tan(\theta_\text{max})]
\]
The lateral acceleration \(a_y(k)\) is the time derivative of the lateral velocity:
\[
a_y = \frac{d}{dt}(v_x \tan(\theta)) \approx v_x \frac{d}{dt}(\tan(\theta))
\]
Assuming \(\omega\) represents the steering rate, we have:
\[
a_y \approx v_x \omega
\]
Therefore, the lateral acceleration constraint is:
\[
a_y(k) \in [-v_x(k)\omega_\text{max}, v_x(k)\omega_\text{max}]
\]
These constraints approximate the nonholonomic constraints of the vehicle by bounding the possible values of lateral velocity and acceleration based on the vehicle's longitudinal velocity and steering capabilities.

Therefore, the vehicle is subject to various constraints:
\begin{subequations}
\begin{align}
    v_y(k) &\in [v_x(k) \tan(\theta_\text{min}), v_x(k) \tan(\theta_\text{max})] \\
    a_y(k) &\in [-v_x(k)\omega_\text{max}, v_x(k)\omega_\text{max}]
\end{align}
\end{subequations}
where the last two constraints approximate the nonholonomic constraints of the vehicle.

\subsection{MIQP Objective Function, Constraints, and Algorithm}
We formulate trajectory planning for autonomous vehicles as a Mixed-Integer Quadratic Programming-based-MPC (MIQP-MPC) \cite{quirynen2024real}. This framework allows for the simultaneous modeling of the vehicle's continuous dynamics and the discrete decision-making processes involved in obstacle avoidance, lane changing, and handling complex road scenarios like speed bumps. The MIQP-MPC provides an effective means of ensuring optimal navigation under such conditions.
The following provides a detailed explanation of each MIOCP formulation as follows,

\begin{subequations}
\begin{align}
    \min_{\mathbf{X}, \mathbf{U}} \quad & \sum_{i=0}^{N} \frac{1}{2} \begin{bmatrix} \mathbf{x}(i) \\ \mathbf{u}(i) \end{bmatrix}^T \mathbf{H}(i) \begin{bmatrix} \mathbf{x}(i) \\ \mathbf{u}(i) \end{bmatrix} + \begin{bmatrix} \mathbf{q}(i) \\ \mathbf{r}(i) \end{bmatrix}^T \begin{bmatrix} \mathbf{x}(i) \\ \mathbf{u}(i) \end{bmatrix} \tag{a} \\
    \text{s.t.} \quad & \mathbf{x}(i+1) = \begin{bmatrix} \mathbf{A}(i) & \mathbf{B}(i) \end{bmatrix} \begin{bmatrix} \mathbf{x}(i) \\ \mathbf{u}(i) \end{bmatrix} + \mathbf{a}(i), \quad \forall i \in \mathbb{Z}_{0}^{N-1}, \tag{b} \\
    & \underline{\mathbf{x}}(i) \leq \mathbf{x}(i) \leq \overline{\mathbf{x}}(i), \quad \underline{\mathbf{u}}(i) \leq \mathbf{u}(i) \leq \overline{\mathbf{u}}(i), \quad \forall i \in \mathbb{Z}_{0}^{N}, \tag{c} \\
    & \underline{\mathbf{c}}(i) \leq \begin{bmatrix} \mathbf{C}(i) & \mathbf{D}(i) \end{bmatrix} \begin{bmatrix} \mathbf{x}(i) \\ \mathbf{u}(i) \end{bmatrix} \leq \overline{\mathbf{c}}(i), \quad \forall i \in \mathbb{Z}_{0}^{N}, \tag{d} \\
    & u_j(i) \in \mathbb{Z}, \quad \forall j \in I(i), \quad \forall i \in \mathbb{Z}_{0}^{N}. \tag{e}
\end{align}
\label{eq:MIOCP}
\end{subequations}

Finally, the MIOCP \eqref{eq:MIOCP} is converted into the following MIQP formulation:

\begin{subequations}
\begin{align}
    \min_{\mathbf{z}} \quad & \frac{1}{2} \mathbf{z}^\top \mathbf{H} \mathbf{z} + \mathbf{h}^\top \mathbf{z} \tag{a} \\
    \text{s.t.} \quad & \mathbf{Gz} \leq \mathbf{g}, \quad \mathbf{Fz} = \mathbf{f}, \tag{a} \\
    & z_j \in \mathbb{Z}, \quad j \in \mathcal{I}, \tag{a}
\end{align}
\end{subequations}

where \(\mathbf{z}\) includes all optimization variables, and the index set \(\mathcal{I}\) denotes the integer variables.

The adoption of Gurobi for solving the MIQP allows our trajectory planning framework to manage the intricate balance between discrete decision-making and continuous optimization, ensuring real-time feasibility in the context of autonomous driving. Its performance is especially beneficial for resource-constrained platforms, enabling our model predictive control (MPC) approach to handle complex dynamic environments while maintaining computational efficiency.
\subsubsection{MIQP Objective Function}
The objective function for the optimal control problem over a prediction horizon of \(N\) steps is given by:
\begin{equation}
\begin{aligned}
    J = \sum_{k=0}^{N-1} \Big( &q_1 (v_x(k) - v_r)^2 + q_2 a_x(k)^2 + q_3 (y(k) - y_r)^2 \\
    &+ q_4 v_y(k)^2 + q_5 a_y(k)^2 + r_1 j_x(k)^2 + r_2 j_y(k)^2 \Big)
\end{aligned}
\label{eq:objective_function}
\end{equation}
where \(q_1, \ldots, q_5\) and \(r_1, r_2\) are weighting coefficients, \(v_r\) is the reference velocity, and \(y_r\) is the reference lateral position. The state variables \(v_x(k)\), \(a_x(k)\), \(y(k)\), \(v_y(k)\), \(a_y(k)\) represent the longitudinal and lateral velocities, accelerations, and positions of the system, while the control inputs \(j_x(k)\) and \(j_y(k)\) represent the longitudinal and lateral jerks at time step \(k\).

To convert the objective function \(J\) into the standard Mixed-Integer Optimal Control Problem (MIOCP) format, we first define the decision vector \(\mathbf{v}(k)\) at each time step \(k\) as:
\[
\mathbf{v}(k) = \begin{bmatrix} x(k) & v_x(k) & a_x(k) & y(k) & v_y(k) & a_y(k) & j_x(k) & j_y(k) \end{bmatrix}^\top
\]
Next, we express the quadratic terms in \eqref{eq:objective_function} in matrix form. The quadratic terms can be rewritten as:
\[
J = \sum_{k=0}^{N-1} \mathbf{v}(k)^\top \mathbf{H} \mathbf{v}(k) + \mathbf{q}^\top \mathbf{v}(k)
\]
where \(\mathbf{H}\) is the matrix containing the quadratic coefficients, and \(\mathbf{q}\) contains the linear terms arising from the reference velocity \(v_r\) and reference lateral position \(y_r\).
The matrix \(\mathbf{H}\) is defined as:
\[
\mathbf{H} = \begin{bmatrix}
0 & 0 & 0 & 0 & 0 & 0 & 0 & 0 \\
0 & q_1 & 0 & 0 & 0 & 0 & 0 & 0 \\
0 & 0 & q_2 & 0 & 0 & 0 & 0 & 0 \\
0 & 0 & 0 & q_3 & 0 & 0 & 0 & 0 \\
0 & 0 & 0 & 0 & q_4 & 0 & 0 & 0 \\
0 & 0 & 0 & 0 & 0 & q_5 & 0 & 0 \\
0 & 0 & 0 & 0 & 0 & 0 & r_1 & 0 \\
0 & 0 & 0 & 0 & 0 & 0 & 0 & r_2
\end{bmatrix}
\]
The vector \(\mathbf{q}\) is given by:
\[
\mathbf{q} = \begin{bmatrix}
0 \\
-2 q_1 v_r \\
0 \\
-2 q_3 y_r \\
0 \\
0 \\
0 \\
0
\end{bmatrix}
\]
Finally, the constant terms (such as \(q_1 v_r^2\) and \(q_3 y_r^2\)) are ignored since they do not affect the optimization problem.
In summary, the objective function \(J\) has been successfully transformed into the quadratic form required for Mixed-Integer Optimal Control Problems, where each term can now be efficiently handled by an appropriate solver:
\[
J = \sum_{k=0}^{N-1} \Bigg[\mathbf{v}(k)^\top \mathbf{H} \mathbf{v}(k) + \mathbf{q}^\top \mathbf{v}(k)\Bigg]
\]

\subsubsection{Speed Bump Negotiation Mimicking Human Driving Behavior Constraints}
In general, the state and control input variables are bounded:
\begin{equation}
\begin{aligned}
    \vect{x}_{\text{min}} \leq &\vect{x}(k) \leq \vect{x}_{\text{max}} \\
    \vect{u}_{\text{min}} \leq &\vect{u}(k) \leq \vect{u}_{\text{max}}
\end{aligned}
\end{equation}

When approaching a speed bump, human drivers typically exhibit certain behaviors to ensure a comfortable and safe passage. Our model aims to mimic these behaviors through a set of logical constraints. We introduce binary variables $\delta_1(k)$, $\delta_2(k)$, and $\delta_3(k)$ to represent the logical conditions related to the speed bump, and $\text{turn\_left}(k)$, $\text{turn\_right}(k)$, and $\text{is\_turning}(k)$ to represent turning behavior. The following constraints implement the logical conditions for speed bump negotiation:

\begin{subequations}
\begin{align}
    &\delta_1(k) = 1 \Leftrightarrow x(k) \geq x_\text{bump\_start} \label{eq:bump1} \\
    &\delta_2(k) = 1 \Leftrightarrow x(k) \leq x_\text{bump\_end} \label{eq:bump2} \\
    &\delta_3(k) = 1 \Leftrightarrow v_x(k) \leq v_\text{max\_bump} \label{eq:bump3} \\
    &\text{turn\_left}(k) = 1 \Leftrightarrow v_y(k) \geq v_\text{turn} \label{eq:turn_left} \\
    &\text{turn\_right}(k) = 1 \Leftrightarrow v_y(k) \leq -v_\text{turn} \label{eq:turn_right} \\
    &\text{is\_turning}(k) = \text{turn\_left}(k) \vee \text{turn\_right}(k) \label{eq:is_turning} \\
    &\delta_1(k) + \delta_2(k) - \text{is\_turning}(k) \leq 1 \label{eq:bump_turn} \\
    &-\delta_1(k) + \delta_3(k) \leq 0 \label{eq:bump_speed1} \\
    &-\delta_2(k) + \delta_3(k) \leq 0 \label{eq:bump_speed2} \\
    &\delta_1(k) + \delta_2(k) - \delta_3(k) \leq 1 \label{eq:bump_speed3}
\end{align}
\end{subequations}

where $x_\text{bump\_start}$ and $x_\text{bump\_end}$ are the start and end positions of the speed bump, $v_\text{max\_bump}$ is the maximum allowed speed over the bump, and $v_\text{turn}$ is the threshold lateral velocity for considering a turn.

By incorporating these human-like behaviors into the optimization model, we aim to generate trajectories that satisfy the mathematical constraints and produce a driving style that feels natural and comfortable to passengers, enhancing the overall experience of autonomous navigation over speed bumps.

\subsubsection{MIQP-MPC Algorithm}
The complete MIQP-MPC algorithm for trajectory planning is summarized as follows:

\begin{algorithm}
\caption{MIQP-MPC for Trajectory Planning}
\begin{algorithmic}
\STATE Initialize state $\mathbf{x}(0)$
\WHILE{not at goal}
    \STATE Formulate MIQP problem with objective function and constraints
    \STATE Solve MIQP to obtain optimal control sequence $\mathbf{u}^*(k:k+N-1)$
    \STATE Apply first control input $\mathbf{u}^*(k)$
    \STATE Update state $\mathbf{x}(k+1)$ based on vehicle dynamics
    \STATE $k \leftarrow k + 1$
\ENDWHILE
\end{algorithmic}
\end{algorithm}

%------------------------------------------------------------------------------
\section{Experimental Results}
\label{sect:Results}
\subsection{Experimental Setup}
To evaluate the proposed method, we implemented the MIQP-MPC algorithm using the Gurobi optimizer. We simulated various driving scenarios on a straight road with a width of 5 meters, including multiple obstacles and a speed bump.
The simulation parameters were set as follows:
\begin{table}[h]
\centering
\begin{tabular}{|l|l|}
\hline
\textbf{Parameter} & \textbf{Value} \\
\hline
\hline
\multicolumn{2}{|c|}{\textbf{Simulation Parameters}} \\
\hline
Prediction horizon (T) & 20 seconds \\
Time step ($\Delta t$) & 0.1 seconds \\
Number of time steps (N) & 200 \\
\hline
\hline
\multicolumn{2}{|c|}{\textbf{Scenario Setup}} \\
\hline
Road width & 2.0 meters \\
Initial x-position ($x_0$) & 0 m \\
Initial y-position ($y_0$) & 0.75 m \\
Initial longitudinal velocity ($v_{x0}$) & 10 m/s \\
Initial lateral velocity ($v_{y0}$) & 0 m/s \\
Reference speed ($v_r$) & 10 m/s \\
Speed bump start ($x_{\text{bump\_start}}$) & 30 m \\
Speed bump end ($x_{\text{bump\_end}}$) & 35 m \\
Maximum speed over bump ($v_{\text{max\_bump}}$) & 5 m/s \\
Wheelbase (L) & 2.7 m \\
Lateral velocity threshold for turning ($v_{\text{turn}}$) & 0.1 m/s \\
\hline
\hline
\multicolumn{2}{|c|}{\textbf{Cost Function Weights}} \\
\hline
$q_1$ (longitudinal velocity deviation) & 1 \\
$q_2$ (longitudinal acceleration) & 1 \\
$q_3$ (lateral position deviation) & 1 \\
$q_4$ (lateral velocity) & 2 \\
$q_5$ (lateral acceleration) & 4 \\
$r_1$ (longitudinal jerk) & 4 \\
$r_2$ (lateral jerk) & 4 \\
\hline
\end{tabular}
\caption{Simulation Parameters, Scenario Setup, and Cost Function Weights}
\label{tab:sim_params_weights_lines}
\end{table}

\subsection{Results and Discussion}
The experiments demonstrated the effectiveness of the proposed MIQP-MPC approach in generating safe and comfortable trajectories for autonomous vehicles in complex scenarios involving obstacles and speed bumps.

\begin{figure}[tb]
	\begin{centering}
	\includegraphics[width=1.0\textwidth]{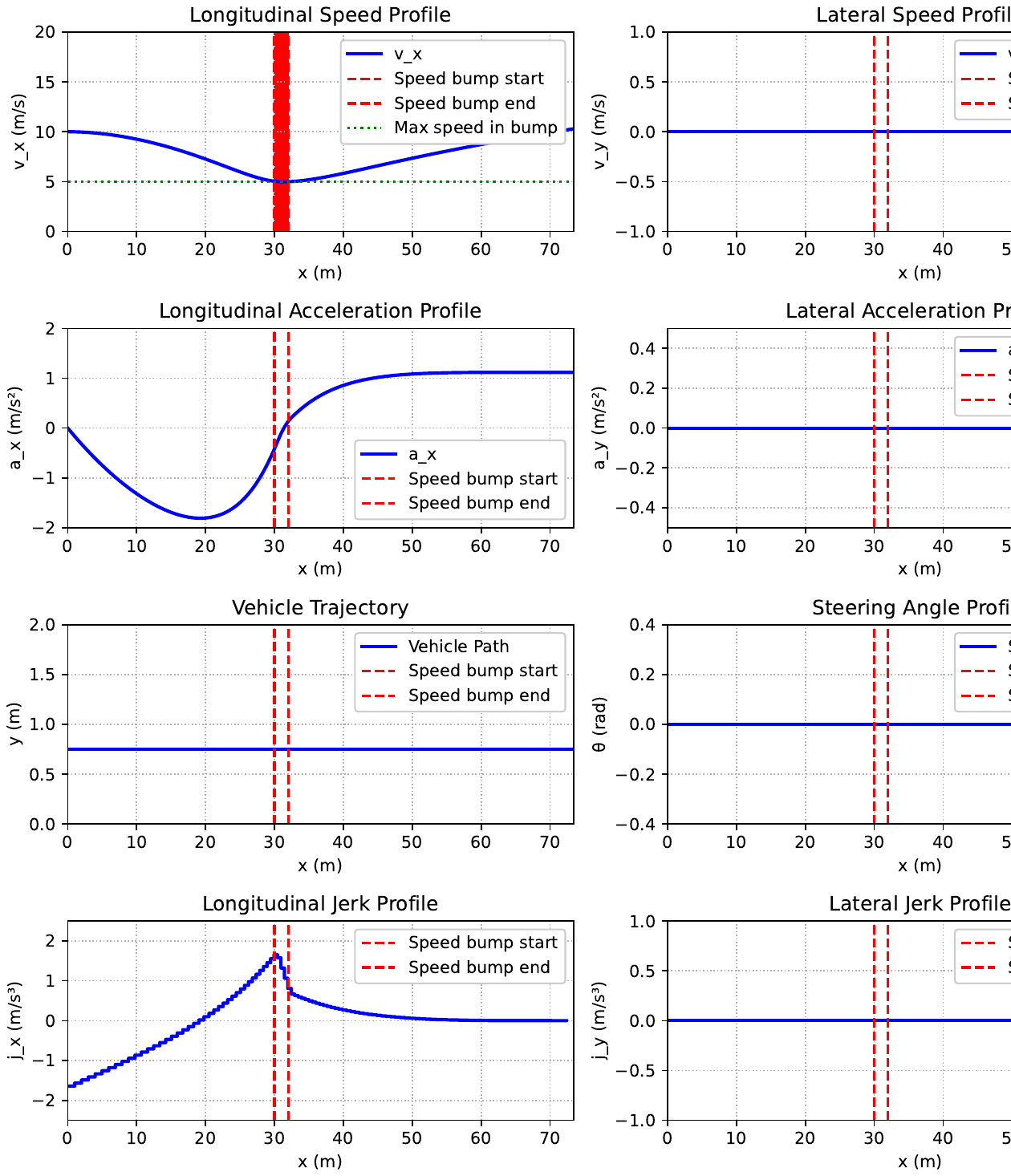}
	\caption{The result with bumper width of 2m- no steering to reduce jerk}
	\label{fig:no_steer}
	\end{centering}
\end{figure}

Fig. \ref{fig:no_steer} illustrates the speed profile during speed bump negotiation. The vehicle consistently reduced its speed when approaching the speed bump, maintaining a speed below the specified limit of 5 m/s while traversing the bump. Note that in this experiment, we do not mimic human driving, and it goes straight to pass the bumper.

\begin{figure}[tb]
	\begin{centering}
	\includegraphics[width=1.0\textwidth]{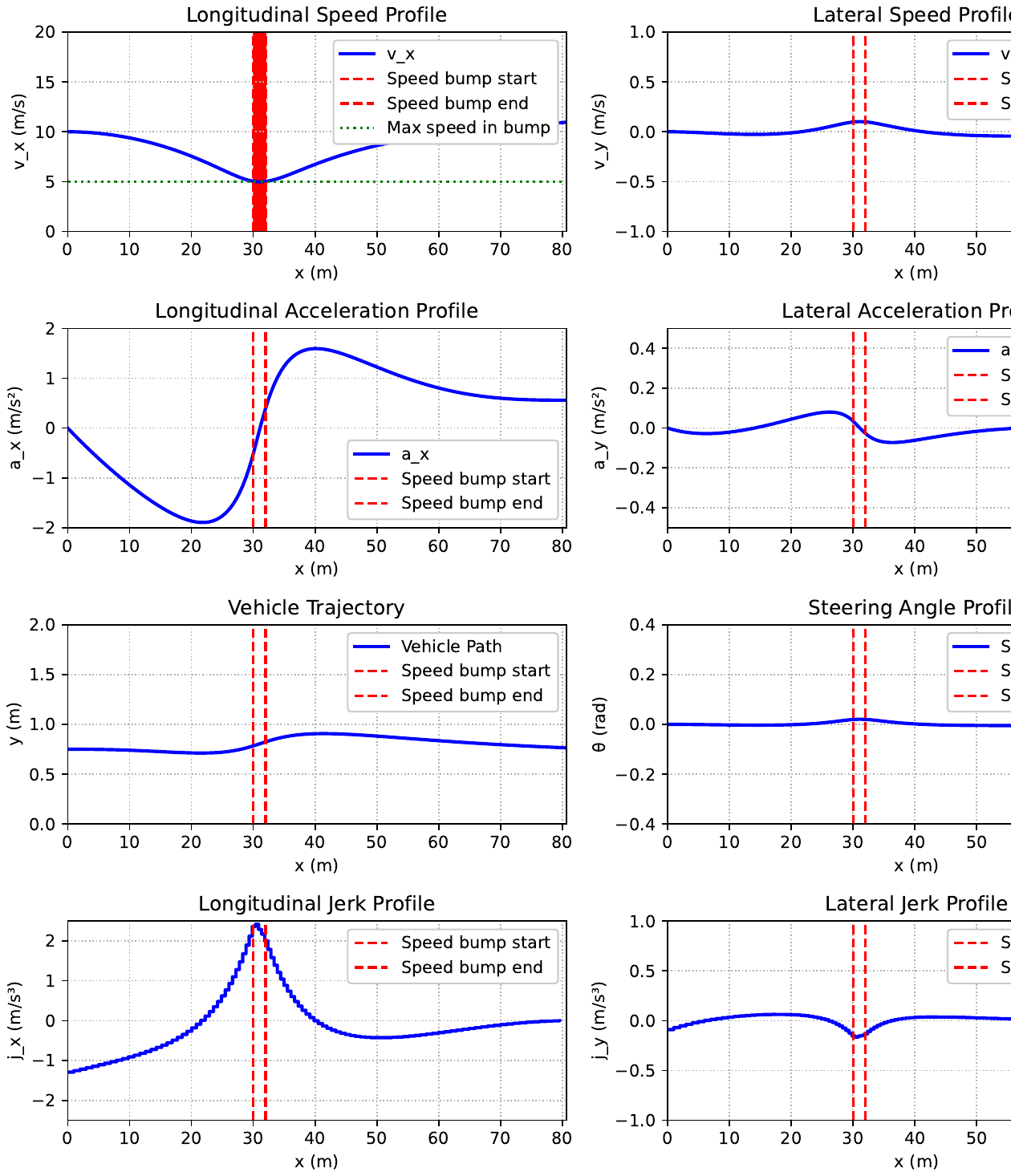}
	\caption{The result with human behavior steering}
	\label{fig:speed_profile}
	\end{centering}
\end{figure}

The subsequent experiment incorporates expert driving behavior, as in fig \ref{fig:speed_profile}. At this time, the system steers the vehicle when it reaches the bumper.

\begin{figure}[tb]
	\begin{centering}
	\includegraphics[width=1.0\textwidth]{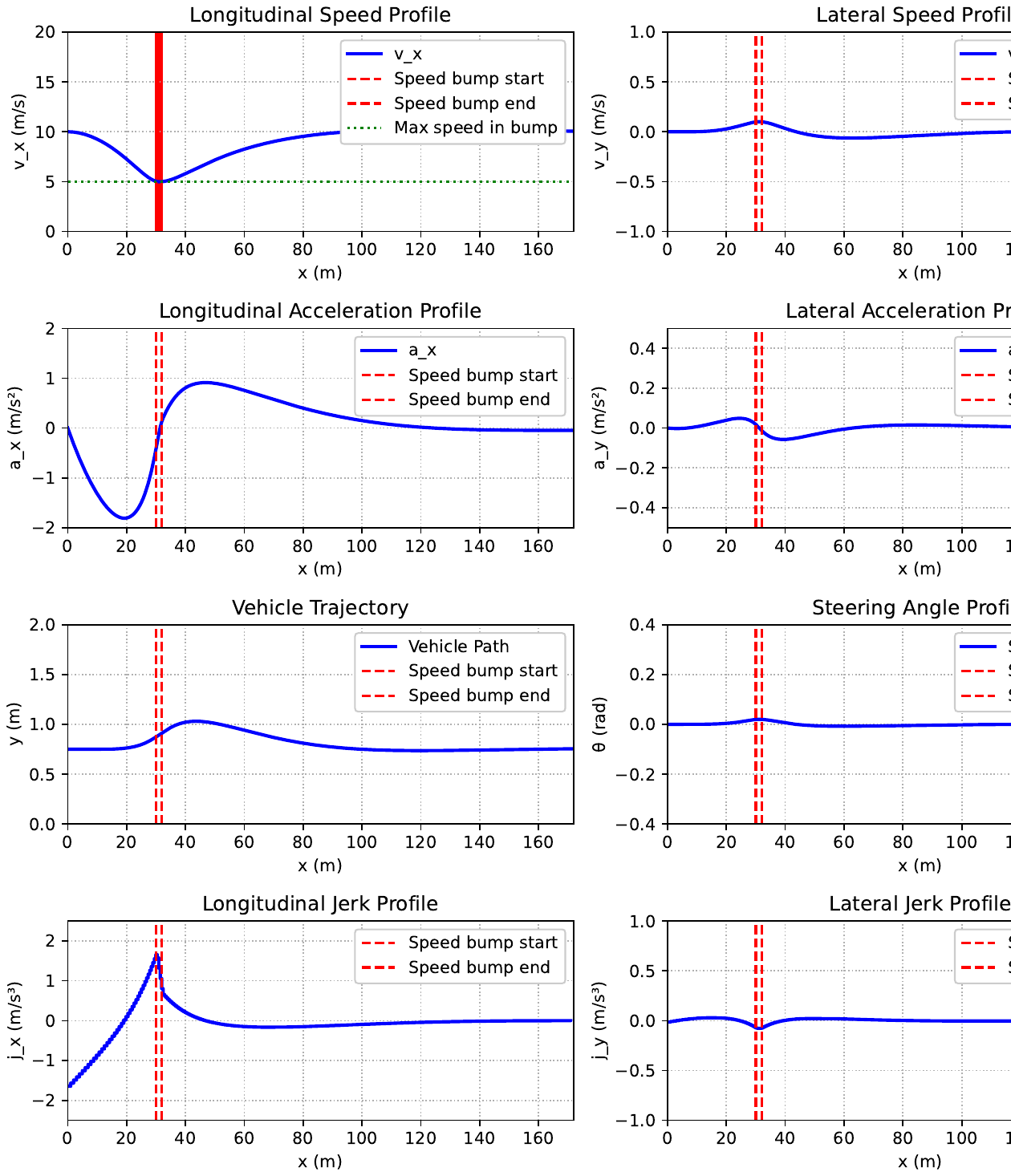}
	\caption{The result with bumper 2m, longer run to converge to reference speed}
	\label{fig:smoothness}
	\end{centering}
\end{figure}

The generated trajectories exhibited smooth changes in position, velocity, and acceleration, minimizing jerk and ensuring passenger comfort. Fig. \ref{fig:smoothness} shows the acceleration and jerk profiles for a typical trajectory.

\begin{figure}[tb]
	\begin{centering}
	\includegraphics[width=1.0\textwidth]{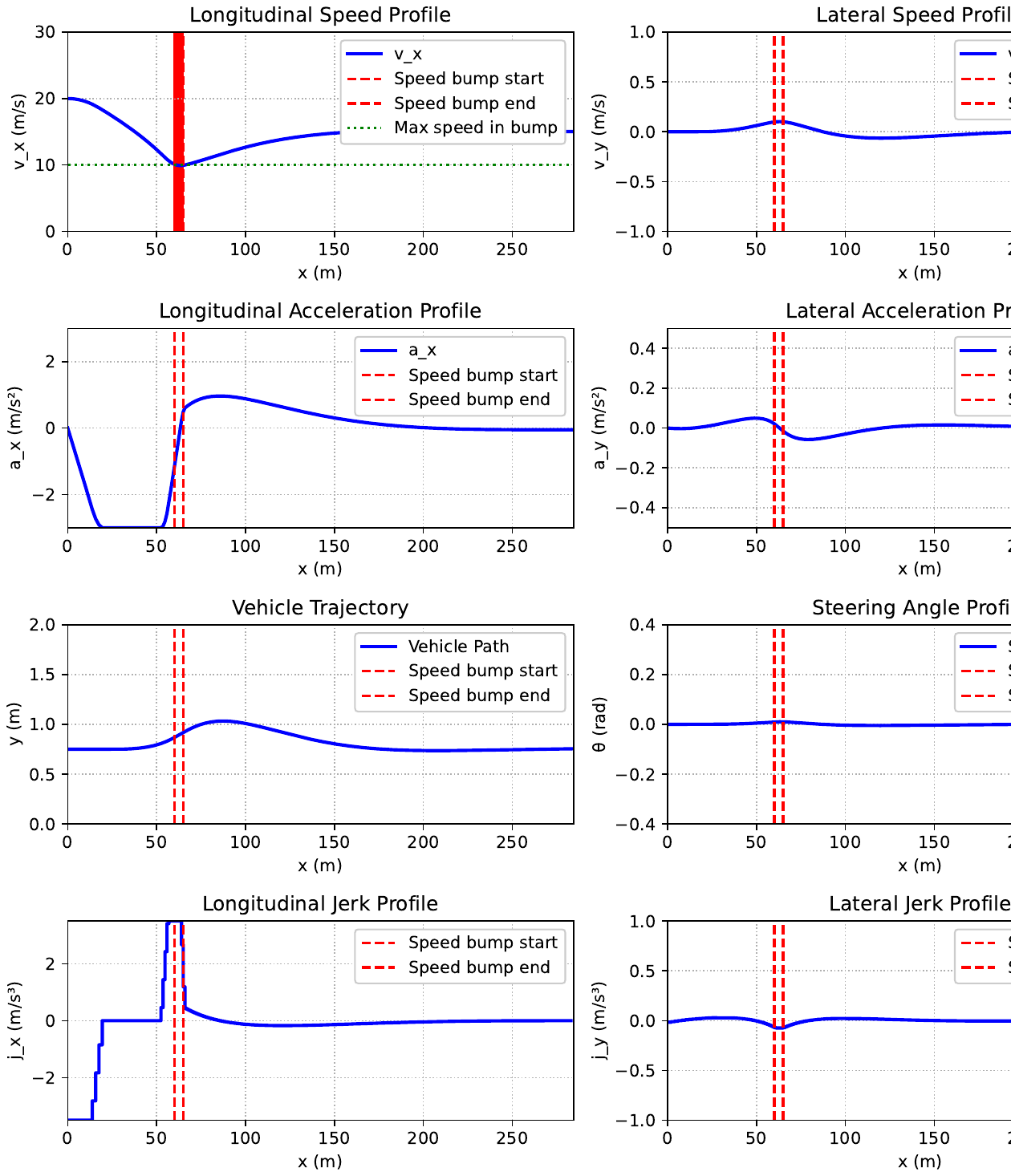}
	\caption{The result handling high-speed over bumper}
	\label{fig:high_speed}
	\end{centering}
\end{figure}

Last but not least, we conducted a study on the high-speed before going to the bumper. The system can easily handle the problem and, after that, track the lateral lane, as shown in Fig. \ref{fig:high_speed}

Despite the computational complexity of MIQP and hardware, our implementation achieves solving times compatible with real-time operation in the tested scenarios. The computational time is approximately 10 ms to 100 ms, and was conducted on a laptop with an Intel Core i5.
However, the current formulation assumes perfect knowledge of obstacle positions and road conditions. Incorporating uncertainty estimation and robust optimization techniques could enhance the method's performance in real-world conditions.

\section{Conclusion}
We propose a new approach to trajectory planning for autonomous vehicles, which incorporates speed bump negotiation using Mixed-Integer Quadratic Programming (MIQP) within a Model Predictive Control (MPC) framework. This method showcases the ability to produce safe, comfortable, and efficient trajectories, particularly in complex urban environments.
The main contributions of this work include: a unified MIQP formulation that simultaneously addresses speed bump negotiation; an MPC-based strategy that generates smooth trajectories while adhering to multiple constraints; and experimental validation demonstrating the method's effectiveness in simulated urban driving conditions.
The results show that the proposed approach navigates complex environments with speed bumps while preserving both passenger comfort and computational efficiency. The ability to handle various constraints within a single optimization framework marks a significant step forward in autonomous vehicle trajectory planning.
In future research, we aim to expand this method to encompass a wider range of urban driving scenarios, including interactions with obstacles, pedestrians, and other vehicles, thereby narrowing the gap between controlled simulations and real-world autonomous driving applications.
%------------------------------------------------------------------------------

\label{sect:bib}
\bibliographystyle{plain}
%\bibliographystyle{alpha}
%\bibliographystyle{unsrt}
%\bibliographystyle{abbrv}
%\bibliography{easychair}

%------------------------------------------------------------------------------
%------------------------------------------------------------------------------
% Index
%\printindex

%------------------------------------------------------------------------------
\end{document}